\documentclass[10pt,twocolumn,letterpaper]{article}

\usepackage{wacv}              % To produce the CAMERA-READY version
\usepackage{graphicx}
\usepackage{amsmath}
\usepackage{amssymb}
\usepackage{booktabs}
\usepackage[accsupp]{axessibility}  % Improves PDF readability for those with disabilities.

\usepackage[pagebackref,breaklinks,colorlinks]{hyperref}

\usepackage[capitalize]{cleveref}
\crefname{section}{Sec.}{Secs.}
\Crefname{section}{Section}{Sections}
\Crefname{table}{Table}{Tables}
\crefname{table}{Tab.}{Tabs.}

\def\wacvPaperID{13} % *** Enter the WACV Paper ID here
\def\confName{WACV}
\def\confYear{2024}

\begin{document}

%%%%%%%%% TITLE - PLEASE UPDATE
\title{NuScenes-MQA: Integrated Evaluation of Captions and QA for Autonomous Driving Datasets using Markup Annotations}

\author{
    Yuichi Inoue, Yuki Yada, Kotaro Tanahashi, Yu Yamaguchi\\
    Turing Inc.\\
    \ \small \{y.inoue, yu.yamaguchi\}@turing-motors.com 
}
\maketitle

%%%%%%%%% ABSTRACT
\begin{abstract}

Visual Question Answering (VQA) is one of the most important tasks in autonomous driving, which requires accurate recognition and complex situation evaluations. However, datasets annotated in a QA format, which guarantees precise language generation and scene recognition from driving scenes, have not been established yet. In this work, we introduce Markup-QA, a novel dataset annotation technique in which QAs are enclosed within markups. This approach facilitates the simultaneous evaluation of a model's capabilities in sentence generation and VQA. Moreover, using this annotation methodology, we designed the NuScenes-MQA dataset. This dataset empowers the development of vision language models, especially for autonomous driving tasks, by focusing on both descriptive capabilities and precise QA. The dataset is available at \hyperlink{https://github.com/turingmotors/NuScenes-MQA}{https://github.com/turingmotors/NuScenes-MQA}.

\end{abstract}

%%%%%%%%%%%%%%%%%%%%%%%%%%%%%%%%%%%%%%%%%%%%%%%%%%%
% Figure
%%%%%%%%%%%%%%%%%%%%%%%%%%%%%%%%%%%%%%%%%%%%%%%%%%%
\begin{figure*}[htb]
\centering
\includegraphics[width=\textwidth]{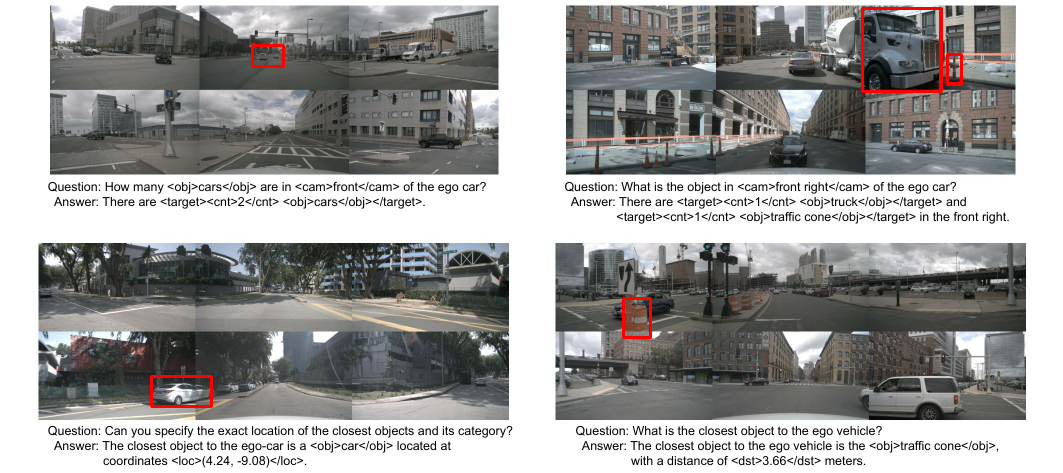}
\caption{Example Scenes and Annotations from NuScenes-MQA Dataset}
\label{fig:sample_annotations}
\end{figure*}

%%%%%%%%% BODY TEXT
%=========================================================================================================================
\section{Introduction}
\label{sec:intro}

Generating accurate text explanations for visual scenes has become crucial in the progress of Vision Language Models (VLMs) integrated with Large Language Models (LLMs). It is believed that many challenges can be more effectively tackled by describing visuals in natural language using VLMs. Specifically, autonomous driving requires accurate recognition and complex situation evaluations. This context underscores the potential benefits of VLMs, which can harness the superior logical reasoning capabilities of LLM. As a result, the pursuit of VLMs tailored for autonomous driving is currently a topic of intense research interest\cite{cui2023survey}.

% \subsection{Autonomous Driving}

The trend of using autonomous driving datasets enriched with natural language annotations has been growing steadily\cite{deruyttere2019talk2car, sha2023languagempc, fu2023drive, xu2023drivegpt4, chen2023driving, mao2023gpt}. By using the data described in natural language, we can align with LLMs that possess general knowledge and high logical reasoning capabilities. By incorporating LLMs, there is potential to create a more intelligent autonomous driving system. To achieve this, it might be necessary to formulate a VLM and curate a relevant dataset for its training. However, datasets annotated in a QA format, which ensures precise language generation and scene recognition from driving scenes, are still a challenge to obtain.

% \subsection{VQA}

Interpreting visual content accurately is not only essential for autonomous driving research, but is also crucial across diverse tasks. Among them, VQA, which provides accurate descriptions of the visual scenes, is particularly important. Various VQA datasets have been proposed, continuously demonstrating their significant value in training and evaluating state-of-the-art VLMs. Traditional QA tasks have predominantly focused on predicting a singular word. However, with the recent proliferation of sophisticated high-performance LLMs\cite{brown2020language}, predicting just one word may not fully harness their potential and may even suppress their inherent linguistic generative abilities. To holistically validate the model's comprehension of the visual content, LLaVA-RLHF\cite{Sun2023-cm} employed RLHF (Reinforcement Learning from Human Feedback) to counteract the vision language model's hallucination. However, this proved to be both time-consuming and costly.

To address this, we introduced "Markup-QA", wherein the QA segment within a naturally composed text is enclosed by our unique markups. Post-processing extracts this markup-wrapped segment, enabling the evaluation of the accuracy of QAs embedded in the text. By removing the markup, the text retains its completeness, allowing us to assess the model's text generation capabilities using standard evaluation metrics. Another advantage of Markup-QA is its flexibility, allowing for the extraction of QAs from any text segment and embedding multiple QAs within a single sentence, marking an innovative departure in QA tasks. 
%Furthermore, adopting this format serves as a simple technique for evaluating the hallucination tendencies of VLM.

Using the rich annotations of nuScenes\cite{Caesar_2020_CVPR} concerning spatial object information, we systematically generated natural language annotations embedded with Markup-QA in a rule-based manner. This dataset, named \textit{NuScenes-MQA}, comprises $1,459,933$ annotations, covering aspects such as object presence, counts, proximity, and relative positions. Using \textit{NuScenes-MQA} ensures simultaneous evaluations of accurate QA capabilities and natural language proficiency.

Our contributions can be summarized as follows:

\begin{itemize}
    \item We introduced Markup-QA, a novel dataset annotation technique in which QAs are enclosed within markups. By using data annotated with Markup-QA, QA tasks can be embedded within natural sentences, allowing a concurrent evaluation of textual quality and QA accuracy. 
    % This significantly contributes to counteracting hallucinations in VLM.
    \item We proposed and publicly released the NuScenes-MQA dataset annotated in the Markup-QA style, along with its evaluation methodology.
    \item Using VLMs capable of handling multiple images, we established a baseline for the NuScenes-MQA dataset.
\end{itemize}

%=========================================================================================================================
\section{Related Work}
\label{sec:formatting}

%-------------------------------------------------------------------------
\subsection{Vision Language Datasets in Driving Scenes}

Datasets for autonomous driving are inherently multimodal, curated from a variety of sensors \cite{Caesar_2020_CVPR, Sun_2020_CVPR, Yu_2020_CVPR, DBLP}. Recently, several existing autonomous driving datasets collected from these sensors have been augmented with textual annotations. For example, the BDD-X\cite{kim2018textual} supplements driving conditions with textual descriptions that explain the underlying reasons. In addition to describing driving scenarios, DriveGPT4\cite{xu2023drivegpt4} uses off-the-shelf object detection models in conjunction with GPT, improving BDD-X with recognition tasks and textual captions. The Honda DRAMA dataset\cite{Malla_2023_WACV} introduces the challenge of localizing risk objects and explicating their risks. In pursuing recognition-specific datasets, NuScenes-QA\cite{qian2023nuscenes} utilizes pre-annotated object information to propose a 3D VQA task. Moreover, DriveLM\cite{drivelm2023} constructs a dataset for nuScenes that encapsulates perception, prediction, and planning, all described in text. The trend underscores the growing attention towards leveraging rich sensor data in autonomous driving to recognize spatial object information and articulate it through natural language.

\subsection{VQA}

The task of VQA involves processing an image and a natural language question to produce a concise natural language response. A wide variety of datasets, such as VQA\cite{Antol_2015_ICCV}, VQA v2.0\cite{Goyal_2017_CVPR}, GOA\cite{Hudson_2019_CVPR}, and Visual Genome\cite{krishna2017visualgenome}, have been introduced. In particular, in the field of autonomous driving, NuScenes-QA\cite{qian2023nuscenes}, which incorporates position information from surrounding objects, is noteworthy. Early research predominantly combined CNN-based image feature extractor with RNNs~\cite{NIPS2016_9dcb88e0, vqa_okatani, Anderson_2018_CVPR}. However, with the emergence of Transformer architecture\cite{NIPS2017_3f5ee243}, transformer-based language models have become the dominant choice for performance. This trend is evident with the introduction of models such as the encoder-decoder-based PALI\cite{ahmadi2023pali, chen2023palix, chen2023pali3}, decoder-only LLaVA\cite{liu2023llava, liu2023improvedllava} and Mini-GPT4\cite{zhu2023minigpt, chen2023minigptv2}, BLIP-2\cite{li2023blip2} with the resampling qformer module, and Flamingo\cite{alayrac2022flamingo} with gated cross-attentions.

\subsection{Special Words in Vision Language Prompts}

Special words improve the efficiency of vision language tasks by better linking visual information to text. The \texttt{<image>} token, for instance, ties image data directly to specific sentences \cite{alayrac2022flamingo, Qwen-VL, peng2023kosmos2}. QwenVL\cite{Qwen-VL} further innovates using tokens such as \texttt{<box>} and \texttt{<ref>} to address the visual grounding task, specifying which textual phrases correspond to the regions of the image. Kosmos-2\cite{peng2023kosmos2} expands on this by associating multiple bounding boxes with phrases using tokens like \texttt{<box>}, \texttt{<delim>}, and location \texttt{<loc>} tokens. These tokens enrich prompts by conveying details not captured by natural language alone.

%%%%%%%%%%%%%%%%%%%%%%%%%%%%%%%%%%%%%%%%%%%%%%%%%%%
% Figure
%%%%%%%%%%%%%%%%%%%%%%%%%%%%%%%%%%%%%%%%%%%%%%%%%%%
\begin{figure*}[htb]
\centering
\includegraphics[width=\textwidth]{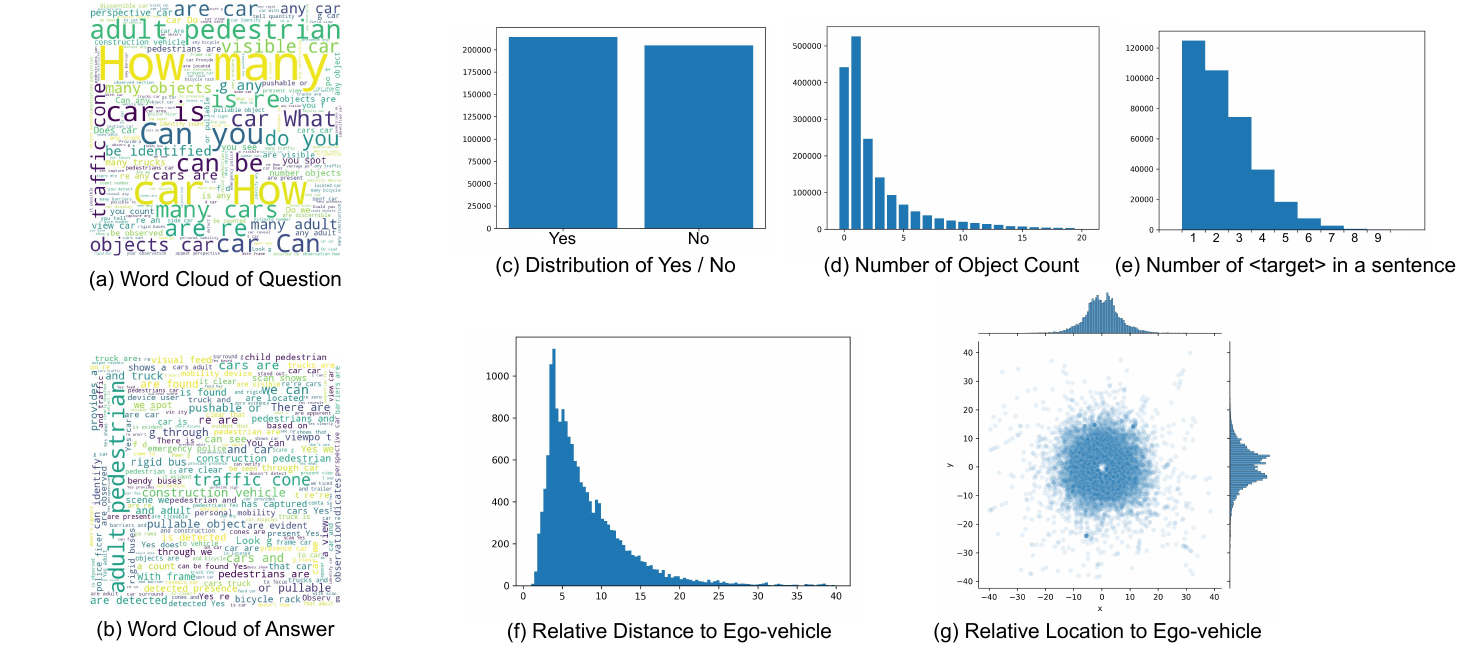}
\caption{Statistics of the NuScenes-MQA dataset}
\label{fig:distribution_qa}
\end{figure*}

% \begin{figure}[htb]
% \centering
% \includegraphics[width=0.45\textwidth]{figures/distribution_simple_qa.png}
% \caption{Simple QA}
% \end{figure}
%------------------------------------------------------------------------

%=============================================
%= Datasetの作成
%=============================================

\section{Proposed Dataset}
\label{sec:formatting}

We introduce a novel QA dataset, called NuScenes-MQA, based on nuScenes\cite{Caesar_2020_CVPR}. Moving away from the conventional short-answer paradigm, our method emphasizes full-sentence responses, enriching both the content and structure of the answers. The dataset targets key aspects of autonomous driving, such as the presence of objects and relative positioning. Using unique markups, we can highlight and evaluate specific information within the answers. With this methodology, we generated $1,459,933$ QA pairs derived from $34,149$ driving scenarios. Example annotations are shown in Fig. \ref{fig:sample_annotations}.

%-------------------------------------------------------------------------
\subsection{Dataset Construction}

In order to create a QA dataset, we employed the annotations provided as ground truth in nuScenes. Contrary to traditional QA datasets that typically structure their response sections with one word, we decided to make our answers as full sentences. To enrich the diversity of our QA templates, we used GPT-4~\cite{openai2023gpt4} and crafted 50 expressions per template, ensuring semantic consistency. Human reviewers then curated and adjusted a subset of 20 to 30 from these generated expressions.

Our dataset is based on four core concepts.

\begin{itemize}
    \item \textbf{Specific Object Presence:} Questions that ask about the existence and number of specific objects.
    \item \textbf{Objects in Specific Direction:} Questions asking for the number and category of objects in a specific direction.
    \item \textbf{Relative Distance to Ego Vehicle:} Questions about the relative distance to vehicles. For simplicity, we identify the object closest to the vehicle and its corresponding distance.
    \item \textbf{Relative location to Ego vehicle:} Questions about the location of objects. Similarly, we simplified the task to identify the closest object and its coordinates.
\end{itemize}

Using the extensive information available in the nuScenes annotations, such as the class and location of recognized objects, and the cameras that capture them, we were able to automate the QA creation process.
%-------------------------------------------------------------------------
\subsection{Markup Implementation}

% This design choice allows for a concurrent assessment of sentence generation accuracy and QA evaluation. To facilitate this, we enveloped the pertinent sections of the QA in markup annotations.

Traditional QA evaluation often revolves around predicting a single word, a method that tends to compromise sentence generation capabilities. To address this, we incorporated special markups in our dataset. These markups were differentiated for each QA type as follows:

\fontsize{10pt}{5pt}\selectfont
\begin{description}
    \item[\texttt{<target>}:]Encapsulates \texttt{<cnt>} and \texttt{<obj>}.
    \item[\texttt{<obj>}:] Represents an object, restricted to a single word.
    \item[\texttt{<cnt>}:] Represents a count, restricted to a single word.
    \item[\texttt{<ans>}:] Represents a binary response, a single word.
    \item[\texttt{<cam>}:] Represents one of the six cameras.
    \item[\texttt{<dst>}:] Represents distance.
    \item[\texttt{<loc>}:] Represents \texttt{(x, y)} coordinates.
\end{description}
\normalsize

By enveloping the target objects with these markups, we can easily extract the relevant words from the answers. For example:

\fontsize{9pt}{9pt}\selectfont
\begin{itemize}
    \item \textit{In the \texttt{<cam>back</cam>}, \texttt{<target><cnt>3</cnt> <obj>trucks</obj></target>are detected.}}
    \item \textit{The closest object to the ego-car is a \texttt{<obj>car</obj>} located at coordinates \texttt{<loc>(3.43, 1.41)</loc>}.}
\end{itemize}
\normalsize

This method allows for the simultaneous evaluation of multiple detections. For example, in the phrase "$3$ trucks", it is necessary to recognize both the class "trucks" and its count "$3$". In this particular regard, conventional methods are insufficient. Using our markup methodology, we can simultaneously evaluate both elements. Similarly, this framework facilitates the recognition of two or more classes at the same time. In the example provided above, it is possible to accurately respond to two distinct questions regarding the object category and the location of the object. Hence, the usage of markups enables the design of tasks that can simultaneously answer multiple queries.

%-------------------------------------------------------------------------
\subsection{Dataset Statistics}

In this section, we discuss the statistics of the NuScenes-MQA dataset. An overview of this dataset is provided through word clouds representing the most frequent terms found within the questions and answers. These visualizations are depicted in Fig. \ref{fig:distribution_qa} (a) and (b), showing the results for the questions and answers, respectively. The word clouds display a particularly diverse range of phrases, especially in the answers.

\subsubsection{Simple QA Task}
\textit{Specific Object Presence} and \textit{Objects in Specific Direction} tasks facilitate recognition-based QA from images, essentially assessing the presence and count of objects. In Fig. \ref{fig:distribution_qa} (c), simple QA tasks seeking binary "Yes" or "No" answers show a slight dominance of "Yes" responses, but the bias does not significantly influence model outcomes. To maintain task feasibility and prevent undue complexity, we excluded questions that required counting more than 20 objects. The distribution of the number of objects to be counted is depicted in Fig. \ref{fig:distribution_qa} (d).

Our dataset harnesses the \texttt{<target>} markup, enabling the embedding of multiple QAs within a single statement. Fig. \ref{fig:distribution_qa} (e) illustrates the number of targets encapsulated in a single statement, highlighting tasks that predict counting and category. Although tasks identifying a single target are common, tasks requiring multiple target responses are also prevalent. This variety allows for an extended evaluation beyond typical simple QA.

\subsubsection{Distance and position distribution}
The nuScenes annotations provide valuable information on the location of objects. Using positional relationships between objects and the ego vehicle, we created QA tasks designated as \textit{Relative Distance to Ego vehicle} and \textit{Relative Location to Ego vehicle}. Drawing inspiration from the range of recognition tasks, such as the occupancy prediction \cite{tian2023occ3d, sima2023_occnet}, we delimited our focus to objects situated within a $40$-meter radius. Fig. \ref{fig:distribution_qa} (f) and (g), which plot the distance and positional relationship of objects, reveal that most are within a $20$-meter range in all directions. To the best of our knowledge, our tasks are the first to require textual answers regarding the spatial information of objects.

%=========================================================================================================================

%%%%%%%%%%%%%%%%%%%%%%%%%%%%%%%%%%%%%%%%%%%%%%%%%%%
% Figure
%%%%%%%%%%%%%%%%%%%%%%%%%%%%%%%%%%%%%%%%%%%%%%%%%%%
\begin{figure*}[htb]
\centering
\includegraphics[width=\textwidth]{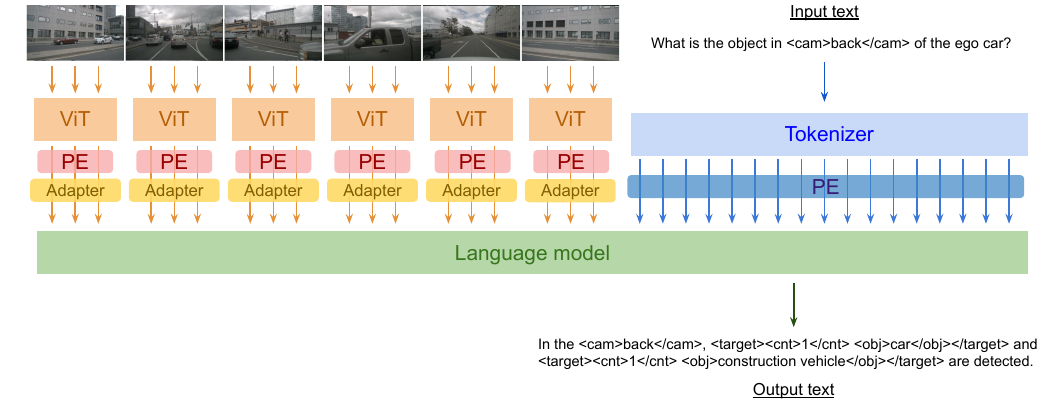}
\caption{Vision Language Model Architecture for NuScenes-MQA}
\label{fig:model_architecture}
\end{figure*}

%------------------------------------------------------------------------
\section{Methodology}
\label{sec:formatting}

\subsection{Model Architecture for Markup-QA}

For our Markup-QA tasks, we introduce a model that combines a vision transformer (ViT) and a decoder-only language model via a simple linear module. The efficiency of this architecture is supported by previous works such as GIT\cite{wang2022git}, LLaVA\cite{liu2023llava, liu2023improvedllava}, and MiniGPTv2\cite{chen2023minigptv2}. The architecture of our model is illustrated in Fig. \ref{fig:model_architecture}.

\begin{figure}[htb]
\centering
\includegraphics[width=0.45\textwidth]{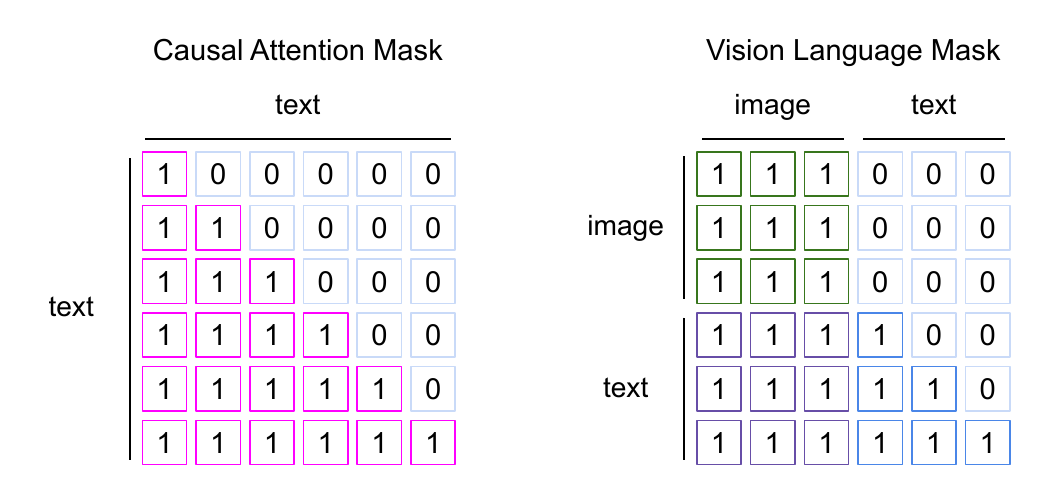}
\caption{Attention Mask for Vision Language Model}
\label{fig:vlm_mask}
\end{figure}

The visual input undergoes feature extraction using a ViT pre-trained in CLIP, subsequently extracting patch features from the final layer. Considering the nuScenes dataset accommodates images from six distinct camera orientations, we use six ViTs, each dedicated to extracting features from its corresponding camera's image. The ViT shares the parameters. These extracted features are subsequently added with six trainable positional embeddings, in a manner similar to \cite{wang2022git}. The ViT patch features are then projected, using a single layer adapter module, to match the size of the text embeddings, thus formulating the visual embeddings. These visual embeddings, similar to text embeddings, are fed into the language model for both training and inference. During training, a causal mask is applied to text embeddings to negate the influence of future information during self-attention computations. However, this constraint is not imposed on visual embeddings, as shown in Fig. \ref{fig:vlm_mask}.

\subsubsection{Tokenization}

Words used for markup adopt a format rarely seen in conventional text. Based on previous studies~\cite{Qwen-VL, peng2023kosmos2}, these markups were typically incorporated as additional tokens. In the results section, we provide a comparison between tokenization using conventional tokenizers and tokenization that treats these markups as additional tokens.

%=========================================================================================================================

%%%%%%%%%%%%%%%%%%%%%%%%%%%%%%%%%%%%%%%%%%%%%%%%%%%
% Figure
%%%%%%%%%%%%%%%%%%%%%%%%%%%%%%%%%%%%%%%%%%%%%%%%%%%
\begin{table*}[htb]
\centering
\includegraphics[width=\textwidth]{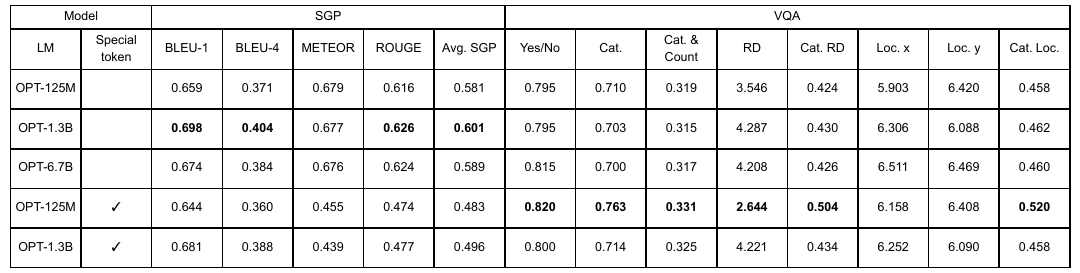}
\caption{Results of different model settings on the NuScenes-MQA test set. We evaluated the natural language generation capability (SGP) and VQA. Cat. means the accuracy of the object category. RD and Loc. represent the relative distance to the ego vehicle and the relative location to the ego vehicle, respectively.}
\label{fig:result_table}
\end{table*}

%------------------------------------------------------------------------
\section{Experiments}
\label{sec:formatting}

\subsection{Evaluation Metrics}

To evaluate the quality of the generated text, we used n-gram-based standard metrics, notably BLEU-1~\cite{papineni2002bleu}, BLEU-4~\cite{papineni2002bleu}, METEOR~\cite{banerjee2005meteor}, and ROGUE-1 F-score~\cite{lin2004rouge}. QA tasks were evaluated in the accuracy metric, while distance measurement tasks were evaluated using the mean absolute error (MAE).

Due to the inherent nature of sentence generation as an evaluation method, inference requires considerable time. To efficiently assess the model's performance, we chose a smaller subset for the model evaluation. From our test set, we carefully extracted $2,000$ samples, ensuring a balanced representation of the tasks. Then, our evaluations were executed on this curated subset.

\subsection{Training Details}

For the ViT, we utilized OpenAI's ViT large patch14, which is pre-trained using CLIP. We used an image resolution of $224$x$224$ during training and evaluation. The language model was OPT~\cite{zhang2022opt}, with parameter sizes of 125M, 1.3B, and 6.7B. Both the ViT and the language model were initialized using pre-trained parameters, while the adapter module started with random initialization. All parameters spanning the ViT, the language model, and the adapter module were trained.

Our training data combined random QAs from a given scene, ensuring that all types of question were present in a sample. The maximum sequence length for the text segment, excluded from visual embeddings, was set at $256$. AdamW optimizer was used, with a learning schedule outlined by a one-cycle scheduler starting at $1e-6$, peaking at $1e-4$, and dropping back to $1e-6$. The other parameters retained their default values. Training was carried out for $10$ epochs using the standard cross-entropy loss. The comprehensive training regimen was orchestrated on $8$ Nvidia A100 GPUs or $8$ Nvidia H100 GPUs.

% Add your bibliography or replace 'ref_opt' with the correct reference tag.

%-------------------------------------------------------------------------
\subsection{Results}

Table~\ref{fig:result_table} shows the performance differences in the Sentence Generation Performance (SGP) and VQA models, according to the size of the model and to the use of markup tokens as additional tokens.

\subsubsection{Sentence Generation Performance}

When evaluating models without the special token, the OPT-1.3B model consistently outperformed its counterparts in most metrics. It achieved the highest scores of BLEU-1, BLEU-4, and ROUGE, with values of $0.698$, $0.404$, and $0.626$, respectively. Although the METEOR scores were comparably high across all three models, the OPT-125M slightly surpassed the others with a score of $0.679$. Furthermore, the OPT-1.3B model achieved a peak \textit{Avg. SGP} score of $0.601$.

In contrast, using markups as a special token led to a marked decline in METEOR scores across all models. Specifically, the OPT-125M model showed the most significant drop, falling to $0.455$. Furthermore, not only METEOR, but other scores also experienced a general decline. Interestingly, for BLEU-1, the OPT-1.3B model with the special token retained a slight advantage, recording a score of $0.681$. The \textit{Avg. SGP} also decreased with the incorporation of the special token. These observations suggest that integrating markup as a special token may adversely impact the language generation capabilities of the models.

\subsubsection{VQA Performance}

Focusing on models without additional tokens, the performance in the \textit{Yes/No} metric remained consistent across all models. In the accuracy of categories, \textit{Cat.}, the OPT-125M model had a slight advantage, achieving a score of $0.710$. Interestingly, for \textit{Loc. x} and \textit{Loc. y}, none of the models showed impressive results, indicating that the task was particularly challenging.

When the special token was integrated, the OPT-125M model showed notable improvements in several metrics such as \textit{Yes/No}, \textit{Cat.}, and \textit{Cat. \& Count}, with scores of $0.820$, $0.763$, and $0.331$, respectively. Both the OPT-125M and OPT-1.3B models showed improvement in the \textit{RD} and \textit{Cat. RD} metrics. Remarkably, the OPT-125M model exhibited significant gains, with the scores for \textit{RD} and \textit{Cat. RD} to $2.644$ and $0.504$, respectively. The OPT-1.3B model exhibited a slight improvement in most of the VQA metrics upon the addition of markup tokens as a special token.

\subsubsection{Quantifying the difficulty of multiple QAs}

In our dataset, following the criteria defined under \textit{Objects in Specific Direction}, a single sentence encompasses multiple QA tasks that require identification of both the categories of objects and their counts. Table~\ref{tab:result_table_nQAs} delineates the accuracy rates based on the number of QAs present in a single sentence (\textit{n}-QA). As the number of QAs increases, tasks become more complex, leading to lower accuracy rates. This decrease is more noticeable in \textit{Cat. \& count} than in \textit{Cat.}. Given these challenges, there is a pressing need for further research to effectively address multiple QAs in a single sentence.

%%%%%%%%%%%%%%%%%%%%%%%%%%%%%%%%%%%%%%%%%%%%%%%%%%%%%%%%%%
%%% Table
%%%%%%%%%%%%%%%%%%%%%%%%%%%%%%%%%%%%%%%%%%%%%%%%%%%%%%%%%%
\begin{table}[h]
\centering
\begin{tabular}{c c c}

\textit{n}-QA & Cat. & Cat. \& Count \\
\hline
\hline
1 & 0.88 & 0.42 \\
2 & 0.59 & 0.16 \\
3 & 0.49 & 0.14 \\
4 & 0.53 & 0.16 \\
5 & 0.51 & 0.13 \\
6 & 0.51 & 0.10 \\
\end{tabular}
\caption{Accuracy based on the number of QAs per sentence}
\label{tab:result_table_nQAs}
\end{table}

%=========================================================================================================================
%------------------------------------------------------------------------
\section{Conclusion}

In this work, we introduced the NuScenes-MQA dataset, which employs a Markup-QA approach where QA is encapsulated within the text using markup. Through the implementation of the Markup-QA scheme, we established a framework that facilitates the simultaneous evaluation of a model's capabilities in sentence generation and VQA. This dataset empowers the development of vision language models, especially for autonomous driving tasks, by focusing on both descriptive capabilities and precise QA. We have also established a baseline model that provides a starting point to demonstrate the practical value of our approach.

\section{Limitation}
While our research offers significant insight, it comes with certain constraints worth noting. First, the dataset has been constructed using a rule-based approach. As a consequence, it may lack the rich diversity often inherent in natural language. This limitation could potentially make it less ideal for training larger models, such as OPT-6.7B, due to potential insufficiency. Furthermore, the limited variety of tasks that address spatial information raises concerns. Specifically, it remains uncertain whether the model truly captures expressions pertaining to positional data. These limitations underscore areas for deeper future research. We believe that future work will overcome these challenges and further advance the field.

%%%%%%%%% REFERENCES
{\small
\bibliographystyle{ieee_fullname}
\bibliography{egbib}
}

\end{document}